\documentclass{article} 
\usepackage{iclr2025_conference,times}
\usepackage{xspace}

\usepackage{hyperref}
\usepackage{url}
\usepackage{graphicx}
\usepackage{subcaption}
\usepackage{multirow}
\usepackage{booktabs}
\usepackage{siunitx}
\usepackage{xcolor}
\usepackage{soul}
\usepackage{pifont}
\usepackage{tcolorbox}
\tcbuselibrary{skins,xparse,breakable}
\tcbset{%
    breakable,
    title filled=false}

\newcommand{\alg}{\texttt{DSMentor}\xspace}

\usepackage{amsmath,amsfonts,bm}

\def\eqref#1{equation~\ref{#1}}

\def\1{\bm{1}}

\DeclareMathAlphabet{\mathsfit}{\encodingdefault}{\sfdefault}{m}{sl}
\SetMathAlphabet{\mathsfit}{bold}{\encodingdefault}{\sfdefault}{bx}{n}

\title{DSMentor: Enhancing Data Science Agents with Curriculum Learning and Online Knowledge Accumulation}

\author{
He Wang\thanks{This work is done when the author is interning in AWS AI.} \\
Carnegie Mellon University \\
\texttt{wanghe@cmu.edu} \\
\AND
Alexander Hanbo Li, Yiqun Hu, Sheng Zhang, Hideo Kobayashi, \\
\textbf{Jiani Zhang, Henry Zhu, Chung-Wei Hang, Patrick Ng} \\
AWS AI \\
\texttt{\{hanboli, yiqunhu, zshe, hideodeo\}@amazon.com} \\
\texttt{\{zhajiani, henghui, cwhang, patricng\}@amazon.com}
}

\definecolor{hew}{RGB}{0,47,167}

\DefineNamedColor{named}{yiqun}{rgb}{0.6,0.2,0.2}

\definecolor{zshe}{RGB}{144,238,144}

\definecolor{todo}{RGB}{255,127,80}

\definecolor{alex}{RGB}{100,200,222}

\iclrfinalcopy 
\begin{document}

\maketitle
\begin{abstract}
Large language model (LLM) agents have shown promising performance in generating code for solving complex data science problems. Recent studies primarily focus on enhancing in-context learning through improved search, sampling, and planning techniques, while overlooking the importance of the order in which problems are tackled during inference. In this work, 
we develop a novel inference-time optimization framework, referred to as \alg, which leverages curriculum learning---a strategy that introduces simpler task first and progressively moves to more complex ones as the learner improves---to enhance LLM agent performance in challenging data science tasks. Our mentor-guided framework 
organizes data science tasks in order of increasing difficulty and incorporates a growing long-term memory to retain prior experiences, guiding the agent's learning progression and enabling more effective utilization of accumulated knowledge.
We evaluate \alg through extensive experiments on DSEval and QRData benchmarks. Experiments show that \alg using Claude-3.5-Sonnet improves the pass rate by up to 5.2\% on DSEval and QRData compared to baseline agents. Furthermore, \alg demonstrates stronger causal reasoning ability, improving the pass rate by 8.8\% on the causality problems compared to GPT-4 using Program-of-Thoughts prompts. Our work underscores the importance of developing effective strategies for accumulating and utilizing knowledge during inference, mirroring the human learning process and opening new avenues for improving LLM performance through curriculum-based inference optimization. 

\end{abstract}

\section{Introduction}\label{sec:introduction}


Data-centric tasks, ranging from statistical analysis to model prediction, are integral to various real-world applications, including healthcare \citep{miotto2018deep}, finance \citep{heaton2017deep}, and engineering \citep{chien2017robotic}. The ever-evolving demands of data-driven fields call for efficient and robust solutions that can effectively process, interpret, and learn from data.

As large language models (LLMs) demonstrate exceptional capabilities in understanding human-like languages, recent works have leveraged LLMs to solve a wide range of tasks, including text generation \citep{dathathri2019plug}, reasoning \citep{wei2022chain}, and code generation \citep{chen2021humaneval}. This has motivated the development of \textit{Data Science agents (DS agents)}---employing LLMs to generate code for data-centric tasks. Thanks to the extensive knowledge built during the training of LLMs, existing DS agents \citep{cheng2023gpt,zhang2023data,zhang2023mlcopilot,hong2024data} have shown strong capabilities in addressing various data science problems.

However, many data science problems are inherently challenging, often involving vague task descriptions, complex data interpretation, and strict output formats, which results in DS agents failing to solve these problems effectively \citep{zhang2024dseval,lai2023ds1000,liu2024qrdata}. Additionally, the knowledge embedded within LLMs is \textit{static},  confined to the data available during their training, posing significant challenges in rapidly evolving fields where up-to-date information is essential for generating accurate and relevant outputs. 

To extend the knowledge of LLMs to specific data-centric tasks, recent works \citep{zhang2023mlcopilot,guo2024ds} primarily focus on retrieving external knowledge for solving data science problems, often starting with identifying or collecting relevant knowledge sources. However, these approaches typically do not accumulate knowledge in a dynamic, online fashion, nor do they explore the importance of task order or curriculum in problem-solving. This oversight is particularly critical in data science, where problems are often interrelated, and complex solutions can frequently be constructed by integrating simpler ones \citep{hong2024data}. For instance, a sophisticated \textit{time series forecasting} task might build upon foundational concepts of \textit{data preprocessing}, \textit{feature engineering}, and basic \textit{regression} techniques. The challenge becomes even more pronounced in online learning or serving scenarios, where the timeliness and quality of the knowledge base directly impact model inference performance. Unfortunately, current methods often result in sub-optimal performance due to inadequate strategies for  how to best prepare, structure, and organize the knowledge base.

\begin{figure}[!t]
    \centering
    \includegraphics[width=\linewidth,trim={3cm 0 3cm 2.5cm},clip]{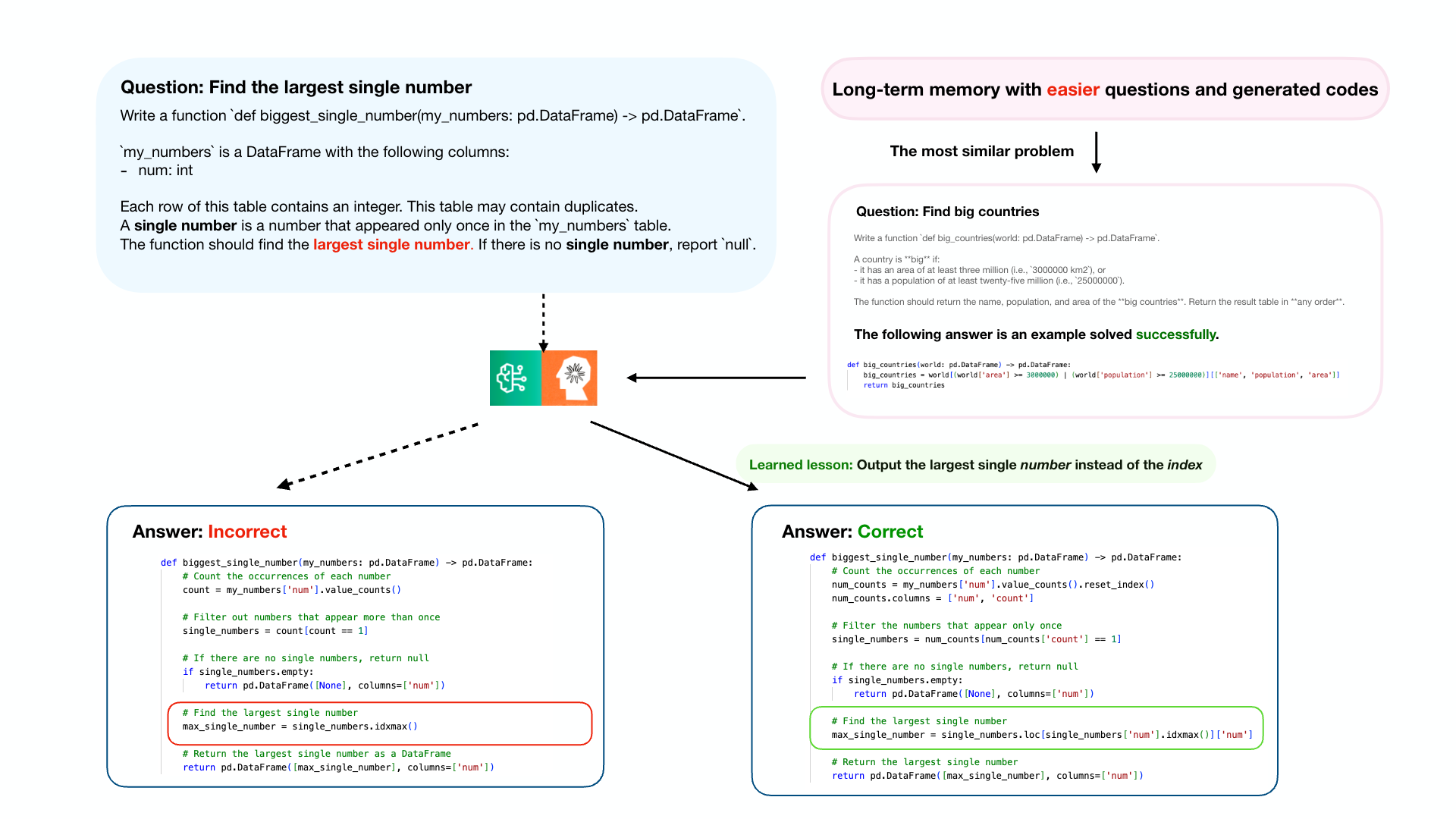}
    \caption{A motivating example illustrating how the agent can learn from previously solved easier problems. Without retrieving information from long-term memory, the agent incorrectly outputs the index of the largest single number, while by leveraging knowledge from previously solved easier questions, the agent can provide the correct answer.}
    \label{fig:motivating example}
    \vspace{-0.8em}
\end{figure}

To address these challenges and leverage the interrelated nature of data science problems, we explore the sequence in which knowledge is populated into and retrieved from a growing memory buffer, helping DS agents better understand and utilize accumulated knowledge. Inspired by the human learning process, where foundational knowledge forms the basis for tackling more complicated problems, we posit that the order in which information is introduced to the model can significantly influence its effectiveness. For instance, understanding basic concepts is often necessary before addressing more advanced challenges, as illustrated in Figure \ref{fig:motivating example}.

In this work, we propose a new framework, referred to as \texttt{DSMentor}, which incrementally enhances the capabilities of DS agents through a mentor-guided approach. The \textit{Mentor} agent evaluates the difficulty of all the tasks and provides a sequence order. Specifically, we employ curriculum learning to carefully structure the learning process for DS agents and accumulate knowledge from the solutions to previous tasks. This approach mirrors the hierarchical nature of data science problem-solving, where complex solutions often build upon simpler concepts. By systematically increasing the difficulty of tasks, our framework enables the agent to develop the skills necessary to tackle more complicated challenges within the data science domain. We evaluate our framework using Claude-3.5-Sonnet and Llama-3.1-70b on two popular data science benchmarks: {DSEval} \citep{zhang2024dseval} and {QRData} \citep{liu2024qrdata}.

Our main contributions can be summarized as follows:
\begin{itemize}
    \item \textbf{Curriculum-based performance improvement:} We demonstrate the superior performance of \texttt{DSMentor}, which leverages an easy-to-hard curriculum, across established benchmarks in data analysis and causal reasoning tasks. Our framework surpasses other data science agents by systematically guiding the learning process, enabling more effective problem-solving, even for complex tasks that require synthesizing advanced knowledge from multiple simpler problems.
    
    \item \textbf{Enhanced knowledge utilization:} We show that retrieving easier and relevant examples from memory and structuring the knowledge in an increasing-similarity order, significantly improves DS agents' understanding and allows them to more efficiently leverage prior experiences for solving new tasks.    
    
    
    \item \textbf{Progressive learning for solving advanced causal problems:} 
    We demonstrate how \texttt{DSMentor} incrementally introduces tasks with increasing complexity, facilitating the agent’s progressive learning of causal relationships. This gradual exposure strengthens its causal reasoning abilities and enables it to effectively address complex data relationships.
\end{itemize}

\section{Related works}

\paragraph{Curriculum learning.}
Inspired by the human learning process, curriculum learning, first introduced by \citep{bengio2009curriculum}, has become a widely-used approach in many applications, including computer vision, natural language processing, and reinforcement learning \citep{wei2016stc, zhang2019leveraging, wang2021survey,portelas2020automatic}. This approach involves organizing training examples in a sequence from easier to harder tasks, which facilitates convergence of the training process and improves the quality of the learned models \citep{hacohen2019power}. Determining task difficulty is a critical factor in enhancing training efficiency \citep{wang2021survey} and can be categorized into pre-defined difficulty \citep{platanios2019competence,shi2023cross,lu2024yoda,liu2024let} and automatic difficulty measurement \citep{portelas2020automatic,wang2023voyager,sun2024conifer}. For instance, the Cross-Episodic Curriculum (CEC) method \citep{shi2023cross} structures learning experiences based on factors such as the pre-defined task difficulty or demonstrator expertise, leveraging cross-episodic attention in Transformers. Recent works \citep{wang2023voyager,sun2024conifer} employ large language models to automatically generate tasks with a diverse range of difficulties. In this paper, we leverage the LLM-based \textit{Mentor} agent to automatically assess task difficulty and systematically construct a curriculum-based dataset that progresses from simpler to more complex tasks. Instead of focusing on the training stage, this paper incorporates curriculum learning with LLM agents during inference, emphasizing the impact of curriculum on the growth of long-term memory and the examples retrieved from online memory.



\paragraph{Data science LLM agents.}
LLMs have become increasingly valuable in data science, including OpenAI's Codex \citep{chen2021evaluating} and Anthropic’s Claude \citep{anthropic2023claude} assisting in data preprocessing, analysis, and code generation through natural language interfaces \citep{biswas2023chatgpt}. However, their static nature, where knowledge is fixed at the time of training, limits their ability to handle real-time data.
To address this, recently proposed agents such as the Data Interpreter \citep{hong2024data}, MLCopilot \citep{zhang2023mlcopilot}, and AIDE \citep{schmidt2024aide} extend LLM capabilities by incorporating dynamic planning, human-like reasoning, and iterative refinement, significantly enhancing performance in data science problem-solving.
Our work builds on these advancements by incorporating curriculum learning and retrieval-based techniques to further enhance LLM agents’ adaptability in data science tasks.

\paragraph{Multi-agent LLM frameworks.}
LLMs have seen growing application in multi-agent systems, where multiple agents collaborate to solve complex tasks through communication and cooperation. Frameworks like CAMEL \citep{li2023camel} and MetaGPT \citep{hong2023metagpt} demonstrate how multi-agent LLM systems can decompose large projects into modular sub-tasks, allowing specialized agents to handle distinct components of the problem. Moreover, curriculum learning and retrieval-augmented techniques are increasingly integrated into these systems to enhance adaptability and efficiency, as shown in recent work focused on game-play agents \citep{wang2023voyager}. Additionally, studies on multi-agent debate systems reveal that agent interaction can significantly improve reasoning abilities  \citep{wang2023apollo, khan2024debating}. In this work, we propose a framework where a \textit{Mentor} agent assists a \textit{Student} agent, which enhances the student agent's problem-solving skills in data science tasks.

\section{DSMentor: a mentor-guided approach}

\begin{figure}
    \centering
    \includegraphics[width=0.95\textwidth,trim={7cm 8cm 7cm 10cm},clip]{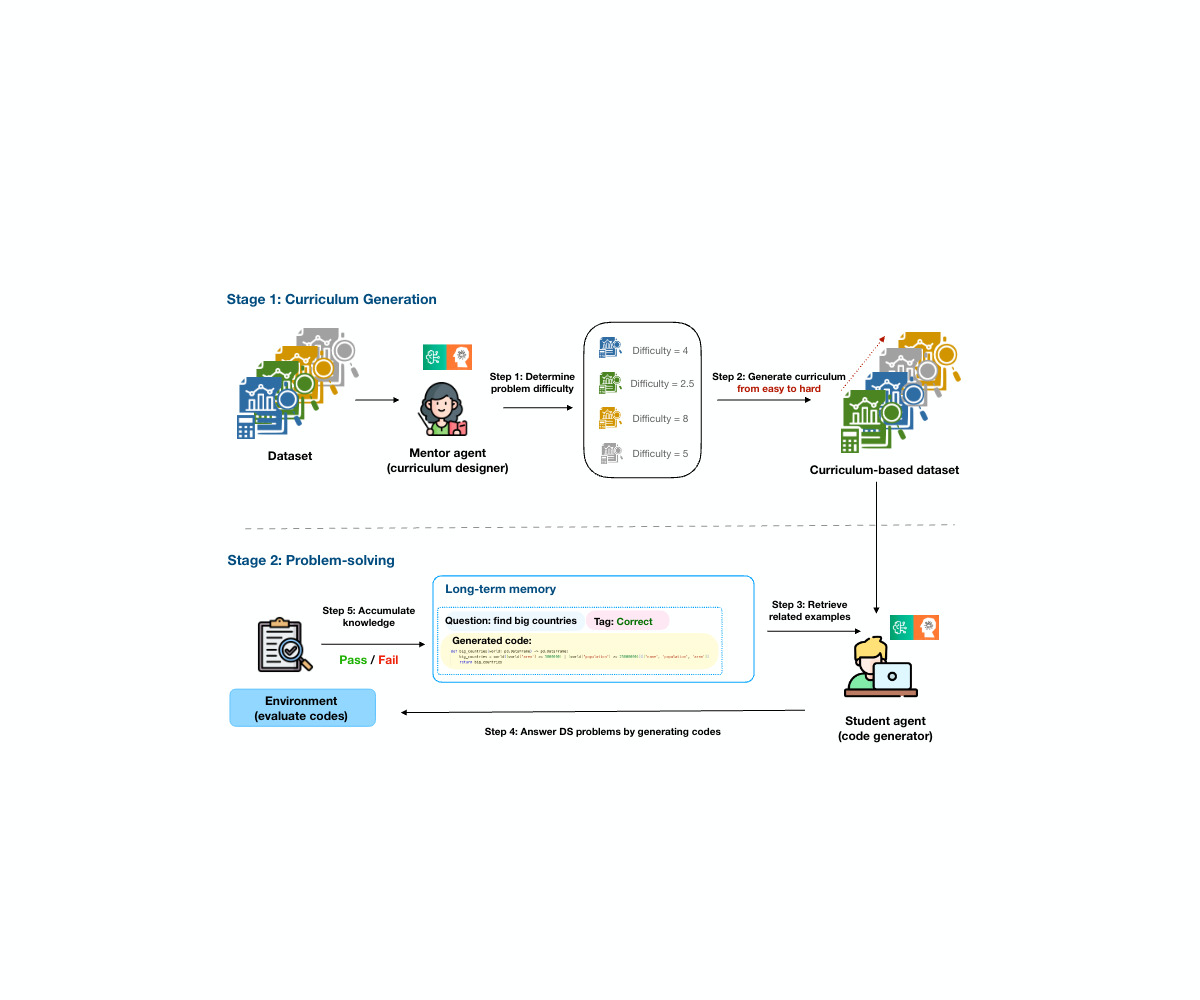}
    \caption{Our framework \alg. Here, the \textit{Mentor} agent assesses the difficulty of each problem and generates a curriculum accordingly. Once the curriculum are determined, the \textit{Student} agent---responsible for answering questions through code generation---retrieves relevant examples from an accumulated online long-term memory. After the environment evaluates the generated code, the \textit{Student} agent will append the question, its output and evaluation tag (i.e., incorrect or correct answer), to the long-term memory.}
    \label{fig:framework}
\end{figure}

In this section, we present \alg, a novel mentor-guided approach that leverages curriculum learning and a growing online memory to enhance the capabilities of DS agents for solving data science tasks. As illustrated in Figure \ref{fig:framework}, \alg operates in two stages: the curriculum-generation stage and the problem-solving stage. In the following subsections, we will discuss the details about these two stages and explain the main components of \alg as well as how they work together to facilitate effective learning.



\subsection{Curriculum-generation stage}
Given any data science task dataset $\mathcal{D}$, we first employ a \textit{Mentor} agent to construct a curriculum-based counterpart $\mathcal{D}_c$. This curriculum establishes the sequence of the tasks that will be undertaken during the problem-solving stage. In this paper, we focus on a difficulty-based curriculum and the preparation of a curriculum-based dataset comprises two key steps.

\begin{figure}[!t]
    \centering
    \includegraphics[width=\linewidth,trim={2cm 3cm 0cm 3.5cm},clip]{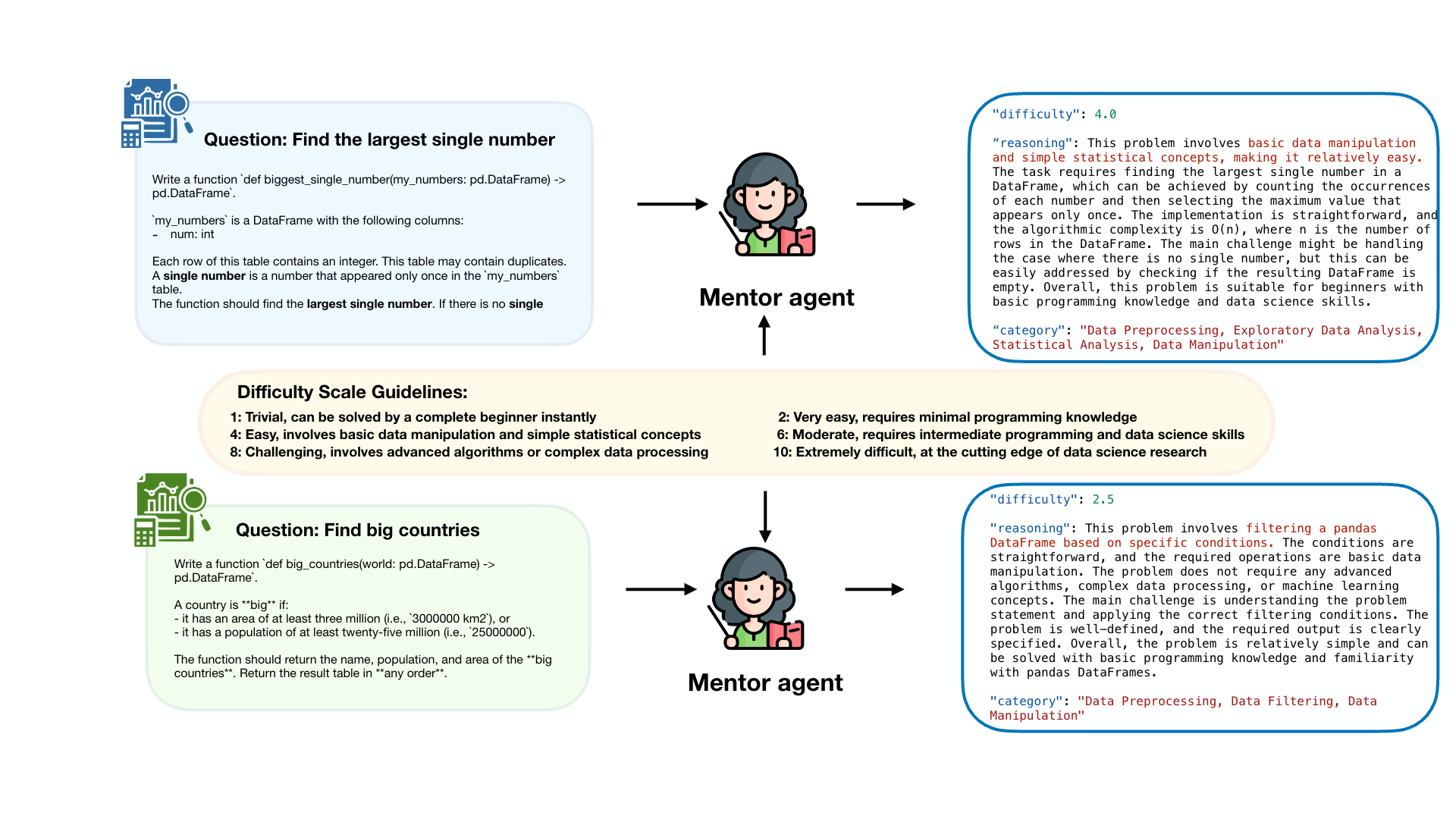}
    \caption{Examples of determining difficulties during the curriculum-generation stage.}
    \label{fig:difficulty_example}
\end{figure}

\paragraph{Step 1: Determine difficulty.} First, the \textit{Mentor} agent assesses and assigns a difficulty level to each problem. In our setting, we rely only on problem descriptions to gauge the difficulty of each problem, since the ground-truth code or evaluation results are generally not available to the \textit{Mentor} agent during this stage in many real world scenarios. To ensure consistency and scalability across a diverse set of DS tasks, we provide \textit{difficulty scale guidelines} for the \textit{Mentor} agent to analyze and assign difficulty levels. As the examples illustrated in Figure \ref{fig:difficulty_example}, the \textit{Mentor} agent can tell that the task of finding the largest single number is more difficult than identifying big counties based on specific conditions. The \textit{Mentor} agent also provides reasons that the former requires counting occurrences within the data, while the latter only involves filtering a DataFrame satisfying the required conditions.

\paragraph{Step 2: Generate curriculum.} Once the difficulty level for each task is determined, we structure the curriculum by arranging the tasks in a sequence that progresses from easier to more challenging problems. Together with the long-term memory described later, this ordering allows the agent to build foundational skills on basic tasks before advancing to more complex ones. For example, as shown in Figure \ref{fig:difficulty_example}, the \textit{Student} agent will first tackle the task of finding big countries, followed by the task of identifying the largest single number, based on their respective difficulty levels.

We also explore the effectiveness of other difficulty metrics and alternative curricula, which will be detailed in \ref{sec:ablation_study} for comprehensive ablation studies.

\subsection{Problem-solving stage}
With the curriculum-based dataset $\mathcal{D}_c$, we utilize a \textit{Student} agent that iteratively solves data science problems, progressing from easier to more complex ones. 
Inspired by verbal reinforcement learning \citep{shinn2024reflexion}, we introduce a growing \textit{online long-term memory} module that enables the \textit{Student} agent to retain and retrieve knowledge from previously tackled, less complex problems, without necessitating updates to the model weights.

\paragraph{Online long-term memory:} Assume that there are $N$ problems in the dataset $\mathcal{D}_c$. For each problem $i\in \{1, \ldots, N\}$, we denote its description as $p_i$, where $i$ is the index of $p_i$ in the sequence of problems. The long-term memory can be formally defined as $$\mathcal{M}_i = \{(p_k,c_k,t_k)\}_{k=1}^{i-1},$$
where $p_k$ and $c_k$ are the problem description and the corresponding generated code of some previous problem $k<i$, and $t_k\in\{\text{Correct},\text{Incorrect}\}$ is the evaluation tag for problem $k$.


\paragraph{Step 3: Retrieve from online long-term memory.}  While solving the $i$-th problem in the curriculum-based dataset $\mathcal{D}_c$, the \textit{Student} agent can retrieve examples from long-term memory $\mathcal{M}_i$ based on the similarity between the current problem description $p_i$ and previous description $p_k\in\mathcal{M}_i$, according to their cosine similarity: $\text{sim}(p_i,p_k) = \cos (\mathcal{E}(p_i),\mathcal{E}(p_k))$, where $\mathcal{E}$ is the embedding model. We finally retrieve examples with the top-$K$ similarities from the long-term memory, where $K$ is the number of the retrieved examples. The examples include both correctly and incorrectly solved problems along with their corresponding generated code.


\paragraph{Step 4: Answer DS problem by generating codes.}
Using the retrieved examples, the \textit{Student} agent follows the order of increasing similarity. By learning from prior experiences, the \textit{Student} agent attempts to solve the current data science problem by generating code, which is then evaluated by the environment based on its execution output.


\paragraph{Step 5: Accumulate knowledge to long-term memory.} After obtaining the evaluation results, the \textit{Student} agent appends the question, the generated code, and the evaluation tag (i.e., correct or incorrect) to the long-term memory for future use. As the inference process continues, the long-term memory will keep expanding.

\section{Experiments}\label{sec:main_experiment}
In this section, we evaluate the proposed \alg on two different data-science benchmarks to address the following research questions: \textbf{(Q1)} How does \alg compare with other popular data science agents? \textbf{(Q2)} Can \alg enhance the ability to solve more difficult problems, like causal reasoning? \textbf{(Q3)} What is most suitable curriculum design that yields the best performance? 

In the sequel, we first attempt to answer \textbf{(Q1)} and \textbf{(Q2)} in Section \ref{sec:main_results} and conduct a series of ablation studies to explore different curriculum designs in Section \ref{sec:ablation_study} for \textbf{(Q3)}.


\subsection{Experiment setup}\label{subsec:experiment_setup}
\paragraph{Benchmarks and datasets.} 
To answer the aforementioned questions, we conducted experiments on {DSEval} \citep{zhang2024dseval} and {QRData} \citep{liu2024qrdata}, which contains 705 problem-sets with 1236 questions in total. More specifically, {DSEval} consists of four datasets: {LeetCode} (with 40 single-turn problem-sets), {StackOverflow} (with 202 single-turn problem-sets), {Exercise} (with 21 multiple-turn problem-sets), and {Kaggle} (with 31 multiple-turn problem-sets), while {QRData} are composed of 411 statistical and causal reasoning problems. More details are summarized in Appendix of the supplementary material, where we address certain DSEval evaluation issues that lead to an underestimation of agent capabilities.  

\paragraph{Implementation details.}

We equip \alg with \textit{anthropic.claude-3-5-sonnet-20240620-v1:0} and \textit{meta.llama3-1-70b-instruct-v1:0} as the base LLMs, referred to as \textit{DSMentor-Claude-3.5-Sonnet} and \textit{DSMentor-Llama-3.1-70b}, respectively. Both the \textit{Mentor} and \textit{Student} agents use the same base model. Moreover, the \textit{Mentor} agent follows the \textit{problem-based difficulty} and the \textit{Student} agent retrieves similar examples from a long-term memory storing previously answered questions and generated codes (\textit{both correct and incorrect}). We employ \textit{cohere.embed-english-v3} as the embedding to retrieve relevant examples from the memory. We set the number of retrieved examples as listed in Appendix of the supplementary material, which will be discussed with in detail in Section \ref{subsec:impact_of_ragnum}. 
During inference, we run \alg with easy-to-hard or hard-to-easy curriculum for three times and \alg with random curriculum for five times to mitigate randomness, reporting average results. The temperature is set to zero for more deterministic behavior.


\paragraph{Evaluation metrics.} For DSEval, we adopt the \textit{Pass Rate} metric from \cite{zhang2024dseval}, which is the ratio of correctly answered questions to the total number of questions, without considering any variable violation issues and error propagation for multiple-turn problem-sets. For QRData, we directly evaluate by comparing the execution output of the generated code against the ground-truth results, which can be either numerical or multiple-choice.

\subsection{Baselines}
We compare \alg to the following baseline agents on DSEval \citep{zhang2024dseval} and QRData \citep{liu2024qrdata}. These agents are selected for their leading performance on the corresponding benchmarks.

\textbf{Jupyter-AI} \citep{jupyterlab2023jupyterai} is an open-source tool that enhances Jupyter Notebooks with generative AI capabilities. It demonstrates strong performance on DSEval-LeetCode, outperforming other data science agent frameworks such as Chapyter \citep{chapyter}, Code Interpreter \citep{codeinterpreter}, and CoML \citep{zhang2023mlcopilot}.

\textbf{CoML \citep{zhang2023mlcopilot}} leverages offline data containing pre-existing tasks and distills the related knowledge to enhance performance during the online stage. It achieves the best performance on DSEval benchmarks, except DSEval-LeetCode compared to other agents \citep{jupyterlab2023jupyterai,chapyter,codeinterpreter}, which makes it an important baseline for comparison.

\textbf{Vanilla agents with Program-of-Thoughts (PoT) \citep{chen2022program}} solve tasks by generating Python code, with the executed output serving as the solution. For each benchmark, Llama-3.1-70b and Claude-3.5-Sonnet act as vanilla agents, using the same system instructions as \alg to generate code (see in the supplementary material). Additionally for QRData, we include GPT-4 \citep{cheng2023gpt} with PoT as one of the baseline agents, as it performs best among PoT-style agents reported in \citet{liu2024qrdata}. To ensure a fair and consistent comparison, we focus on single-turn code generation alternatives rather than ReAct-style \citep{yao2022react} or other agents that involve multi-turn reasoning and error feedback.





\subsection{Results and Discussion}\label{sec:main_results}



\begin{table}[!t]
\centering
\small
\scalebox{0.9}{
\begin{tabular}{l S[table-format=1.3] S[table-format=1.3] S[table-format=1.3] S[table-format=1.3]}
\toprule
\textbf{Model (with Llama-3.1-70b)} & \textbf{DSEval-LeetCode} & \textbf{DSEval-SO} & \textbf{DSEval-Exercise}& \textbf{DSEval-Kaggle}  \\
\midrule
Jupyter-AI \citep{jupyterlab2023jupyterai} &  0.733 & 0.427 & 0.679 &  0.465\\
CoML \citep{zhang2023mlcopilot}        &  0.683& 0.794& \textbf{0.783}& \textbf{0.577}\\
Llama-3.1-70b (PoT)                        & 0.725 & 0.739 & 0.747 & 0.539 \\
\textbf{DSMentor-Llama-3.1-70b }                        & \textbf{0.792} & \textbf{0.837}& 0.752 &  \textbf{0.577}\\
\midrule
\textbf{Model (with Claude-3.5-Sonnet)} & \textbf{DSEval-LeetCode} & \textbf{DSEval-SO} & \textbf{DSEval-Exercise}& \textbf{DSEval-Kaggle}  \\
\midrule
Jupyter-AI \citep{jupyterlab2023jupyterai} &  0.850 & 0.477 & 0.693 & 0.595 \\
CoML \citep{zhang2023mlcopilot}        &  0.875& 0.809& 0.643&  0.535\\
Claude-3.5-Sonnet (PoT)                      & 0.800 & 0.804 & 0.745 & 0.641\\
\textbf{DSMentor-Claude-3.5-Sonnet}                 & \textbf{0.892} & \textbf{0.837}  & \textbf{0.781 }& \textbf{0.678}\\
\bottomrule
\end{tabular}}
\caption{Performance comparison of our \alg and existing data science models across DSEval-LeetCode, DSEval-SO, DSEval-Kaggle, and DSEval-Exercise.}
\label{tab:framework-comparison-dseval}
\end{table}

\begin{table}[!t]
\centering
\small
\scalebox{0.9}{
\begin{tabular}{l c c c}
\toprule
\textbf{Model} & \textbf{Pass Rate} & \textbf{Multiple Choice/Numerical} & \textbf{Statistical/Causal} \\
\midrule
GPT-4 (PoT) & 0.491 &  0.460/\textbf{0.540} & \textbf{0.725}/0.368 \\ 
Llama-3.1-70b (PoT)                       & 0.442 & 0.480/0.384  & 0.615/0.351\\
Claude-3.5-Sonnet (PoT)                      & 0.471 & 0.495/0.436  & 0.646/0.379\\
\midrule
\textbf{DSMentor-Llama-3.1-70b} & 0.508  & 0.566/0.419 & 0.676/\textbf{0.419} \\
\textbf{DSMentor-Claude-3.5-Sonnet} & \textbf{0.543} & \textbf{0.602}/0.452 &  0.707/\textbf{0.456} \\
\bottomrule
\end{tabular}}
\caption{Performance comparison of our \alg and existing data science models on QRData, which includes the overall pass rate, along with breakdowns for multiple choice/numerical and statistical/causal questions.}
\label{tab:framework-comparison-qrdata}
\end{table}

\paragraph{Baselines vs DSMentor.} As shown in Table  \ref{tab:framework-comparison-dseval} and \ref{tab:framework-comparison-qrdata}, DSMentor-Claude-3.5-Sonnet consistently outperforms baseline agents across multiple datasets. Compared to prior art Jupyter-AI and CoML, it achieves significant improvements, including an \textbf{8.8\% increase} on  DSEval-Exercise and and \textbf{8.3\% boost} on DSEval-Kaggle. DSMentor-Llama-3.1-70b also shows notable gains, with an \textbf{5.9\% improvement} on DSEval-LeetCode and a \textbf{4.3\% improvement} on DSEval-SO. Despite performing slightly worse on DSEval-Exercise compared to CoML with Llama-3.1-70b,  \alg demonstrates strong performance on QRData as shwon in Table \ref{tab:framework-comparison-qrdata}, which contains larger and more difficulty problem sets.

\paragraph{Online long-term memory.} Compared to the vanilla Llama-3.1-70b and Claude-3.5-Sonnet using PoT prompts, \alg, equipped with the same LLM, significantly improves the pass rate on both DSEval and QRData. Such improvements are driven by the presence of online long-term memory and curriculum learning. Furthermore, they are especially noticeable on more difficult datasets, such as DSEval-Kaggle and QRData, where leveraging previous examples plays a crucial role in enhancing performance.



\paragraph{Causal reasoning tasks.} 
Causal reasoning is considered more challenging than other statistical problems in QRData \citep{liu2024qrdata}. From Table \ref{tab:framework-comparison-qrdata}, DSMentor-Claude-3.5-Sonnet achieves 45.6\%, which improves  GPT-4 (PoT) as the best baseline model by 8.8\% on the causal problems. Such an improvement indicates that accumulating knowledge from easier questions is particularly effective for complex reasoning, especially in cases where understanding causal relationships is crucial.

In summary, \alg, incorporating with curriculum learning and long-term memory, shows a noticeable advantage over baseline models, particularly on complex datasets like DSEval and QRData. Our results underscore the importance of scalable curriculum strategies and long-term memory in improving AI-assisted data science performance.

\subsection{Ablations}\label{sec:ablation_study}

Next, we conduct a series of ablation studies to examine the impacts of each components of \alg, regarding curriculum and long-term memory designs.

\subsubsection{Difficulty definition} \label{subsec:impact_of_difficulty_definition}

\begin{table}[!t]
\centering
\small
\scalebox{0.9}{ 
\begin{tabular}{l S[table-format=1.3] S[table-format=1.3]}
\toprule
\textbf{Difficulty Measurement} & \textbf{DSEval-LeetCode} & \textbf{DSEval-SO} \\
\midrule
Manual                    & 0.742 & {--} \\
Reference-code-based       & 0.750 & 0.812 \\
Pass-rate (Llama-3.1-8b)   & \textbf{0.800} & 0.825 \\
Pass-rate (Claude-3-Haiku) & 0.775 & 0.812 \\
\midrule
Problem-based only   & 0.792 & \textbf{0.837} \\
\midrule
\textit{baseline w/o curriculum:} Llama-3.1-70b (PoT)           & 0.725 & 0.739 \\
\bottomrule
\end{tabular}}
\caption{Comparison among different difficulty designs for DSMentor-Llama-3.1-70b, where the numbers of retrieved examples are 5 and 15 for DSEval-LeetCode and DSEval-SO respectively. We also report baseline numbers without curriculum in the last row.} 
\label{tab:difficulty-definition_llama}
\end{table}

\begin{table}[!t]
\centering
\small
\scalebox{0.9}{ 
\begin{tabular}{l S[table-format=1.3] S[table-format=1.3]}
\toprule
\textbf{Difficulty Measurement} & \textbf{DSEval-LeetCode} & \textbf{DSEval-SO} \\
\midrule
Manual                      &  0.892&  {--}  \\
Reference-code-based        & 0.825 & 0.832\\
Pass-rate (Llama-3.1-8b)    & 0.883& 0.827\\
Pass-rate (Claude-3-Haiku) & 0.867 & 0.815\\
\midrule
Problem-based only   &  \textbf{0.892 }  &  \textbf{0.837}\\
\midrule
\textit{baseline w/o curriculum:} Claude-Sonnet-3.5 (PoT)        & 0.800 & 0.804\\
\bottomrule
\end{tabular}}
\caption{Comparison among different difficulty designs for DSMentor-Claude-3.5-Sonnet, where the number of retrieved examples is 5 for both DSEval-LeetCode and DSEval-SO.}
\label{tab:difficulty-definition_sonnet}
\end{table}

Figuring out an appropriate difficulty measurement plays a critical role in our framework. In addition to the problem-based difficulty used in Section \ref{sec:main_experiment}, where the LLM-based Mentor agent assesses difficulty based solely on each question, we evaluate three additional difficulty metrics and analyze their performance on DSEval-LeetCode and DSEval-SO. Below, we summarize the key concepts of these difficulty metrics, where more details are postponed to Appendix.

\textbf{Manual difficulty (for DSEval-LeetCode only):} 
referring to the difficulty level (easy/medium/hard) as defined on LeetCode. Problems are first sorted by their original difficulty level, and in case of a tie, human pass rate is used to further rank them (i.e., a higher pass rate indicates an easier problem).

\textbf{Reference-code-based difficulty:} following the difficulty in DSEval \citep{zhang2024dseval}, based on the number of functions, variables, conditions, and loops in  the reference code.

\textbf{Pass-rate difficulty:} employing the pass rate of weaker models, such as \textit{Llama-3.1-8b} and \textit{Claude-3-Haiku},  where a higher pass rate indicates an easier question.

To explore the impact of different difficulty definitions, we follow the same experiment setup described in Section \ref{subsec:experiment_setup} except the curriculum generation process that utilizes different difficulty metrics. We run each experiment for three times and present the average result in Table \ref{tab:difficulty-definition_llama} and \ref{tab:difficulty-definition_sonnet} to mitigate the randomness. As shown in Table \ref{tab:difficulty-definition_llama} and \ref{tab:difficulty-definition_sonnet}, we first observe that incorporating with the curriculum designs significantly improves the overall pass rate for both DSMentor-Llama-3.1-70b and  DSMentor-Claude-3.5-Sonnet on DSEval-LeetCode and SO, comparing to the baseline vanilla agent without memory and curriculum. Moreover, the problem-based difficulty generally outperforms the other difficulty metrics, although DSMentor-Llama-3.1-70b using the problem-based difficulty performs slightly worse than when using the pass-rate generated by Llama-3.1-8B on DSEval-LeetCode. We further examine the correlation between problem-based difficulty and pass-rate difficulty that exclusively reflects each model's problem-solving abilities for each tasks. Due to page limits, we postpone the detailed discussion and results to Appendix.

\subsubsection{Order of tasks and examples}
In addition to difficulty metrics, we further explore the impact of different task orders and retrieved examples, both of which are key components of curriculum design.

First, we compare the easy-to-hard curriculum with the random and hard-to-easy curriculum, where the retrieved examples are ordered by increasing similarity (denoted as \textit{Inc. Similarity}). As shown in Table \ref{tab:curriculum-comparison-llama} and \ref{tab:curriculum-comparison-sonnet}, both DSMentor-Llama-3.1-70b and DSMentor-Claude-3.5-Sonnet perform better with the easy-to-hard or hard-to-easy curricula compared to the random curriculum, demonstrating the efficacy of curriculum learning. Additionally, the easy-to-hard curriculum often shows better performance than the hard-to-easy curriculum, except for DSEval-Kaggle, where the hard-to-easy curriculum performs slightly better by 0.4\%. 

Given that curriculum design influences performance through retrieved examples, we also investigate different ranking approaches for these examples. In addition to increasing similarity, we use increasing difficulty (denoted as \textit{Inc. Difficulty}) for easy-to-hard curriculum. As seen in Table \ref{tab:curriculum-comparison-llama} and \ref{tab:curriculum-comparison-sonnet}, our experimental results show that the easy-to-hard curriculum with examples ordered by increasing similarity often outperforms alternative approaches, except on DSEval-Kaggle.

\begin{table}[!t]
\centering
\small
\scalebox{0.9}{
\begin{tabular}{l S[table-format=1.3] S[table-format=1.3] S[table-format=1.3] S[table-format=1.3]  S[table-format=1.3]}
\toprule
\textbf{Curriculum Design} & \textbf{DSEval-LeetCode} & \textbf{DSEval-SO}& \textbf{DSEval-Exercise} & \textbf{DSEval-Kaggle} \\
\midrule
Easy-to-Hard (Inc. Similarity)  & \textbf{0.792 }& \textbf{0.837}& \textbf{0.752}&  0.577 \\
Easy-to-Hard (Inc. Difficulty)  & 0.775 & \textbf{0.837}& 0.729&  0.566 \\ 
\midrule
Hard-to-Easy (Inc. Similarity)  & \textbf{0.792 }&  0.807 &0.740& \textbf{ 0.583} \\
\midrule
Random (Inc. Similarity)        & 0.770& 0.803 &  0.750     & 0.571\\
\bottomrule
\end{tabular}}
\caption{Performance comparison of DSMentor-Llama-3.1-70b with different curriculum designs across DSEval. Here, Inc. Similarity and Inc. Difficulty represent that the retrieved examples are in the order of increasing similarity or increasing difficulty, respectively.}
\label{tab:curriculum-comparison-llama}
\end{table}

\begin{table}[!t]
\centering
\small
\scalebox{0.9}{
\begin{tabular}{l S[table-format=1.3] S[table-format=1.3] S[table-format=1.3] S[table-format=1.3]  S[table-format=1.3]}
\toprule
\textbf{Curriculum Design} & \textbf{DSEval-LeetCode} & \textbf{DSEval-SO}& \textbf{DSEval-Exercise} & \textbf{DSEval-Kaggle} \\
\midrule
Easy-to-Hard (Inc. Similarity)  & \textbf{ 0.892 }& \textbf{0.837}&\textbf{0.781} &0.678 \\
Easy-to-Hard (Inc. Difficulty)  & 0.850& 0.835  & 0.774 & \textbf{0.684}\\ 
\midrule
Hard-to-Easy (Inc. Similarity)  & 0.833 & 0.797 & 0.742 & 0.680\\
\midrule
Random (Inc. Similarity)        & 0.850 &0.800 &  0.775 &0.678 \\
\bottomrule
\end{tabular}}
\caption{Performance comparison of DSMentor-Claude-3.5-Sonnet with different curriculum designs across DSEval.}
\label{tab:curriculum-comparison-sonnet}
\end{table}



\subsubsection{Long-term memory}
As the \textit{Student} agent utilizes examples retrieved from long-term memory, we then examine the impact of long-term memory, specifically focusing on the number of retrieved examples and the inclusion of incorrectly answered question with failed attempts.

\begin{figure}[!t]
    \centering
    \includegraphics[width=0.9\textwidth]{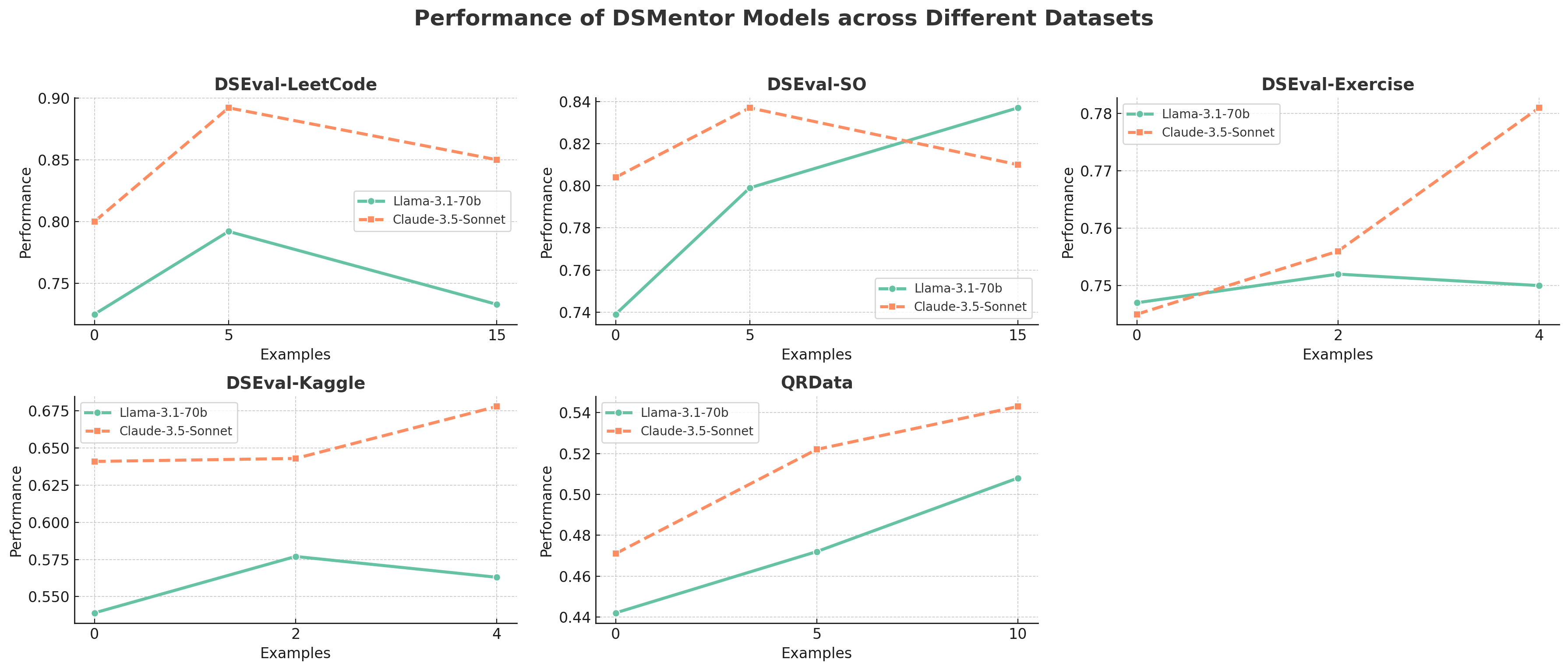}
    \caption{Performance of DSMentor models across different datasets on DSEval and QRData, with varying number of retrieved examples. The subfigures show the results for each dataset, demonstrating performance trends for Llama-3.1-70b and Claude-3.5-Sonnet models.}
    \label{fig:rag_num}
\end{figure}

\paragraph{The impact of the number of retrieved examples.}\label{subsec:impact_of_ragnum}
We first vary the number of retrieved examples and examine its impact for both DSMentor-Llama-3.1-70b and DSMentor-Claude-3.5-Sonnet over all the five datasets. Except the number of examples, all setups follow Section \ref{subsec:experiment_setup}. From Figure \ref{fig:rag_num}, \alg with retrieved examples (i.e. \# examples $> 0$) consistently improves performance across all datasets. However, we observe that DSMentor-Llama-3.1-70b often experiences a performance drop as the number of retrieved examples increases, particularly on multi-turn datasets such as DSEval-Exercise and DSEval-Kaggle. In contrast, DSMentor-Claude-3.5-Sonnet generally benefits from additional retrieved examples, often demonstrating improved performance with more retrieved examples.



\vspace{-0.4em}

\paragraph{The impact of adding incorrect examples.}
Next, we further explore whether the \textit{Student} agent could benefit from incorrect examples (i.e., previously answered questions that are incorrect). We evaluate DSMentor-Llama-3.1-70b on DSEval and QRData, both with and without incorrect examples retrieved from the long-term memory. Apart from incorporating incorrect examples, all experimental setups follows Section \ref{subsec:experiment_setup}. Table \ref{tab:wrong_memory} shows that enabling access to incorrect examples can enhance the pass rate, especially for datasets with fewer problem-sets (e.g., DSEval-LeetCode and DSEval-Exercise), though the datasets with a larger number of problem-sets (e.g., DSEval-SO, DSEval-Kaggle and QRData), the improvement from adding incorrect examples is marginal. This observation aligns with the principle that larger datasets generally possess a more extensive memory pool for retrieval, thereby increasing the likelihood of selecting sufficient correct examples without necessitating additional learning signals from incorrect instances.
\vspace{-0.2em}
\begin{table}[!t]
\centering
\small
\scalebox{0.9}{ 
\begin{tabular}{>{\centering\arraybackslash}m{3.5cm} c S[table-format=1.3] S[table-format=1.3]}
\toprule
\textbf{Dataset} & \textbf{Incorrect Examples} & \textbf{Pass Rate} \\
\midrule
\multirow{2}{*}{DSEval-LeetCode} & \ding{55} & 0.733 \\
                                 & \ding{51} & \textbf{0.792} \\
\midrule
\multirow{2}{*}{DSEval-SO}       & \ding{55} & 0.830 \\
                                 & \ding{51} & \textbf{0.837} \\
\midrule
\multirow{2}{*}{DSEval-Exercise} & \ding{55} & 0.725 \\
                                 & \ding{51} & \textbf{0.752} \\
\midrule
\multirow{2}{*}{DSEval-Kaggle}   & \ding{55} & 0.562 \\
                                 & \ding{51} & \textbf{0.577} \\
\midrule
\multirow{2}{*}{QRData}          & \ding{55} & 0.475 \\
                                 & \ding{51} & \textbf{0.508} \\
\bottomrule
\end{tabular}}
\caption{DSMentor-Llama-3.1-70b with and without incorrectly answered examples.}
\label{tab:wrong_memory}
\end{table}
\vspace{-0.2em}



\section{Conclusion}

In this work, we develop \alg, a framework that enhances data science agents through a mentor-guided curriculum learning approach, effectively improving their ability to tackle complex data science tasks. Extensive experiments demonstrate that organizing tasks from easy to hard significantly boosts the agent's problem-solving and causal reasoning capabilities by building a strong knowledge foundation during inference. Future directions include exploring adaptive curriculum that adjust task difficulty based on agent performance,  incorporating advanced memory mechanisms for better knowledge retention, and extending \alg to support multi-agent collaboration in complex, interdisciplinary domains.

\bibliography{ref}
\bibliographystyle{iclr2025_conference}


\end{document}